\documentclass[letterpaper, 10 pt, conference]{ieeeconf}  

\IEEEoverridecommandlockouts                              

\usepackage[utf8]{inputenc} 
\usepackage[T1]{fontenc}    
\usepackage{lineno, color}
\usepackage{url}            
\usepackage{booktabs}       
\usepackage{amsfonts}       
\usepackage{nicefrac}       
\usepackage{microtype}      
\usepackage{graphicx}
\usepackage{wrapfig}
\usepackage{amsmath}
\usepackage{algorithm}
\usepackage{algpseudocode}
\usepackage{multirow}
\usepackage{hyperref}
\hypersetup{
  colorlinks = true, 
  urlcolor   = magenta, 
  linkcolor  = blue, 
  citecolor  = red 
}

\usepackage[capitalize]{cleveref}
\crefname{section}{Sec.}{Secs.}
\Crefname{section}{Section}{Sections}
\Crefname{table}{Table}{Tables}
\crefname{table}{Tab.}{Tabs.}

\title{\LARGE \bf
Scalable Multi-Robot Collaboration with Large Language Models: Centralized or Decentralized Systems?
}

\author{Yongchao Chen$^{1,2}$, Jacob Arkin$^{1}$, Yang Zhang$^{3}$, Nicholas Roy$^{1}$, and Chuchu Fan$^{1}$
\thanks{$^{1}$Massachusetts Institute of Technology. jarkin@mit.edu, nickroy@csail.mit.edu, chuchu@mit.edu }
\thanks{$^{2}$Harvard University. yongchaochen@fas.harvard.edu}
\thanks{$^{3}$MIT-IBM Watson AI Lab. yang.zhang2@ibm.com}
}

\begin{document}

\maketitle
\thispagestyle{empty}
\pagestyle{empty}

\begin{abstract}
A flurry of recent work has demonstrated that pre-trained large language models (LLMs) can be effective task planners for a variety of single-robot tasks. The planning performance of LLMs is significantly improved via prompting techniques, such as in-context learning or re-prompting with state feedback, placing new importance on the token budget for the context window. An under-explored but natural next direction is to investigate LLMs as multi-robot task planners. However, long-horizon, heterogeneous multi-robot planning introduces new challenges of coordination while also pushing up against the limits of context window length. It is therefore critical to find token-efficient LLM planning frameworks that are also able to reason about the complexities of multi-robot coordination. In this work, we compare the task success rate and token efficiency of four multi-agent communication frameworks (centralized, decentralized, and two hybrid) as applied to four coordination-dependent multi-agent 2D task scenarios for increasing numbers of agents. We find that a hybrid framework achieves better task success rates across all four tasks and scales better to more agents. We further demonstrate the hybrid frameworks in 3D simulations where the vision-to-text problem and dynamical errors are considered. See our project website\footnote[4]{https://yongchao98.github.io/MIT-REALM-Multi-Robot/\label{website}} for prompts, videos, and code.
\end{abstract}

\section{INTRODUCTION}
Multi-robot systems have great potential as a tool for operations that require the completion of many tasks, such as warehouse management. Planning for these systems is often challenging due to heterogeneous robot capabilities, coordination during tasks requiring multiple robots, inter-dependencies of separate tasks, and general safety considerations (e.g. collision avoidance). Further, the difficulty scales with the number of robots. Previous work has used algorithm-based \cite{Catl-for-multi-agent,autotamp,CBF-for-multi-agent} or learning-based \cite{RL-for-multi-agent-coordination,starcraft} methods to control multi-robot systems. Such approaches are typically tuned for a specific scenario, requiring significant engineering effort that limits generalization into novel tasks or scenarios.

Motivated by the ability of pre-trained large language models (LLMs) to generalize to new task domains \cite{llms-few-shot-learners,llms-zero-shot-reasoners}, there have been many recent efforts to use them for single-agent task planning \cite{saycan,llms-zero-shot-planners}. Planning performance is significantly improved through clever use of the context provided to the LLM, whether via techniques for initial prompts (e.g., in-context learning, chain-of-thought) or iterative re-prompting with feedback (e.g., environment state changes, detected errors). Given this success, there is new interest in investigating LLMs as task planners for multi-robot systems \cite{yilun-multi-agent,roco}. These recent efforts address systems consisting of two or three robots; they assign an LLM to each robot and have the models engage in collaborative dialogue rounds to try to find good plans.

Scaling to systems of many robots and tasks with longer horizons is an issue for approaches that assign each robot its own LLM agent. First, both the number of possible coordinating actions and the possible action inter-dependencies grow exponentially with the number of agents, making the reasoning more difficult for the language models. Second, the context provided to each LLM contains the responses of each other LLM for the current round of dialogue in addition to the history of dialogue, actions, and states from prior rounds; so, scaling the number of agents also scales the context token length requirements toward their modern limits and increases the runtime of LLM inference (and API costs). Moreover, the immediately relevant information in the context can become diluted in longer prompts. These limitations are beyond the scope of prior work \cite{yilun-multi-agent,roco}.

\begin{figure}[ht]
  \centering
  \includegraphics[width=0.9\linewidth]{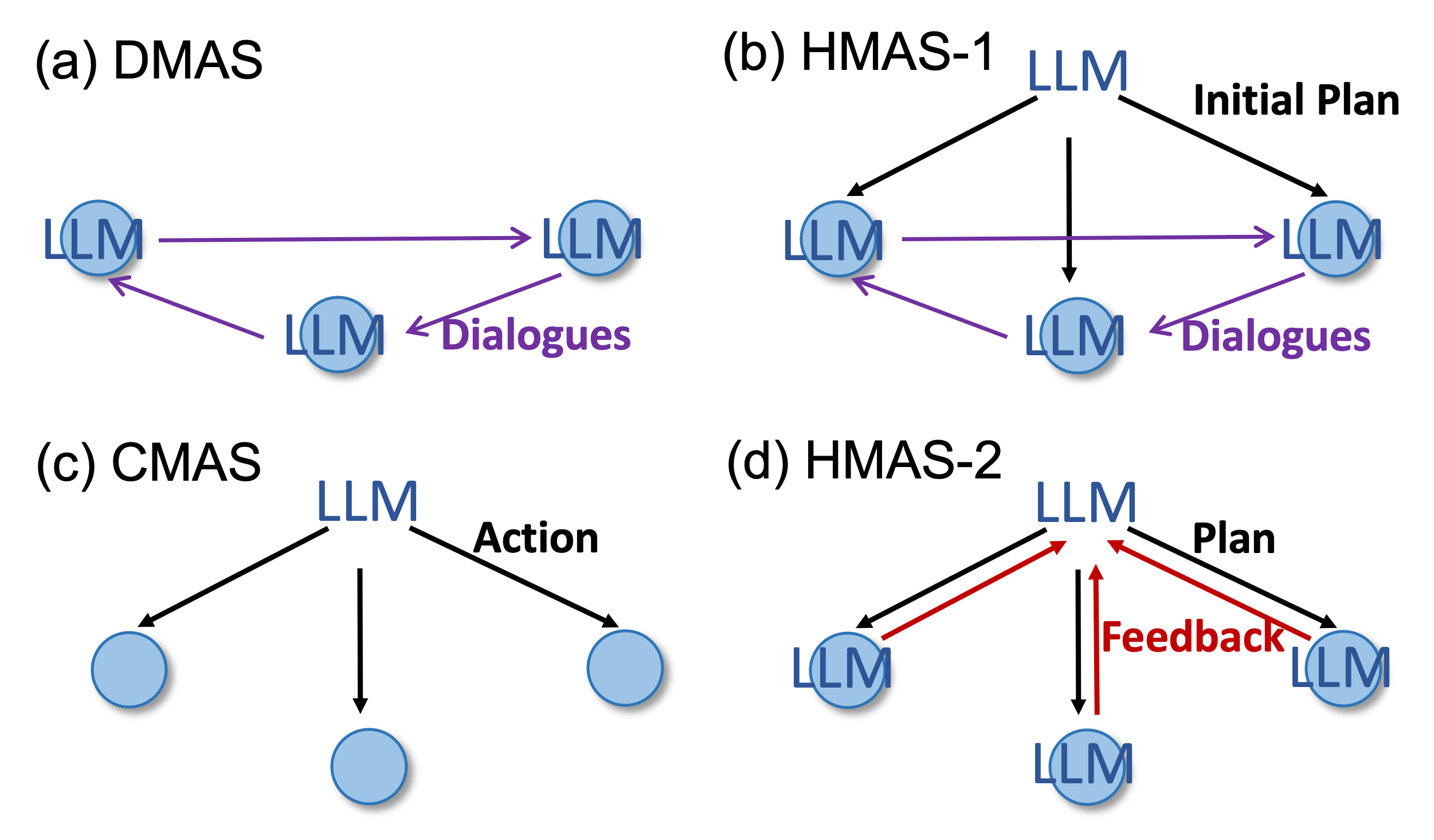}
   \caption{
Four LLM-based multi-agent communication frameworks compared in this work. The circles represent robots that may have actions in the current step and the 'LLM' text represents each LLM agent. The overlap between one circle and one 'LLM' text means that the robot is delegated with one LLM agent to express its special opinions to other agents. The 'LLM' text without the overlapped circle represents a central planning agent.}
   \label{fig:All frameworks}
\end{figure}

Our goal is to preserve the generalizability of LLMs as task planners for multi-robot settings while addressing the challenges of scaling to many agents. We argue that different frameworks for integrating LLM planners into multi-robot task planning can improve both scalability and task planning success rates. In this work, we compare four different frameworks (Figure \ref{fig:All frameworks}) of cooperative dialogue for task planning among multiple LLMs for increasing numbers of robots. For each, planning is performed incrementally in which the LLMs collaborate to find the next action to take for each agent in the system. The first approach (DMAS) uses a decentralized communication framework in which each robot is provided its own LLM agent and dialogue proceeds in rounds of turn-taking. The second approach (CMAS) uses a centralized framework in which a single LLM produces the next action for all robots in the system. We also propose two hybrid versions of these two approaches: (1) a variant of DMAS that adds a central LLM responsible for providing an initial plan to prime the dialogue (HMAS-1) and (2) a variant of CMAS that gives each robot an LLM with which to provide robot-local feedback to the central LLM planner. To further address issues of token length due to historical dialogue and planning context, we also propose a truncated prompt that only includes state-action information from prior dialogue rounds. We evaluate the performance of each approach in four different task planning environments inspired by warehouse settings. To further demonstrate LLMs as multi-robot planners, we apply these approaches to a simulated 3D manipulation task that requires coordination among the manipulators.

To the best of our knowledge, this work is the first to study the scalability of different LLM dialogue frameworks for multi-robot planning. The proposed HMAS-2 framework and state-action history method greatly improve the planning success rates for situations under high robot numbers.

\section{PROBLEM DESCRIPTION}
This work focuses on task planning for multi-robot systems. We consider a cooperative multi-robot task scenario with $N$ robots and $M$ LLM agents. We assume that each LLM agent has full knowledge of the environment and each robot's capabilities. The robot capabilities can be heterogeneous, requiring the planners to assign tasks to robots accordingly. In order to provide each LLM with the task goals and observations, we manually define functions to translate them into text prompts. We also define functions to map the output of the LLM planners into pre-defined robot actions. Planning is performed iteratively, choosing the next action for each robot to take. At each iteration, the $M$ LLM agents engage in collaborative dialogue to find a consensus for the next set of robot actions. Given the next action, the robots act in the environment, and the resulting new state is provided as context to the LLMs for the next planning iteration.

\begin{figure}[ht]
  \centering
  \includegraphics[width=1.0\linewidth]{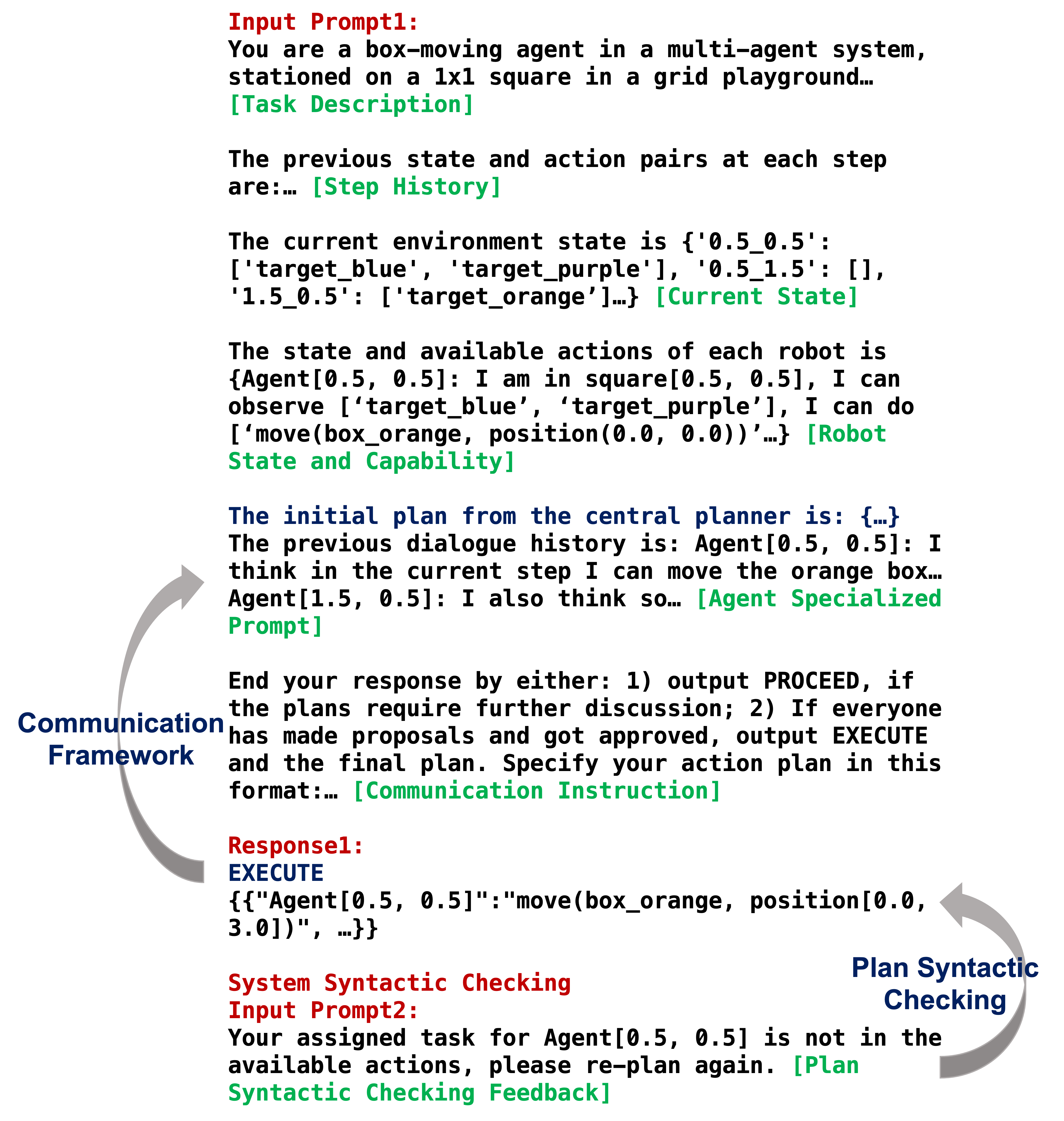}
   \caption{
Simplified prompt example of the HMAS-1 local agent. The acquired 'Response1' acts as the initial plan, or otherwise sent to the next local agent for further discussion.}
   \label{fig:Prompt example simplified version2}
\end{figure}

\begin{figure}[ht]
  \vspace{0.5mm}
  \centering
  \includegraphics[width=1.0\linewidth]{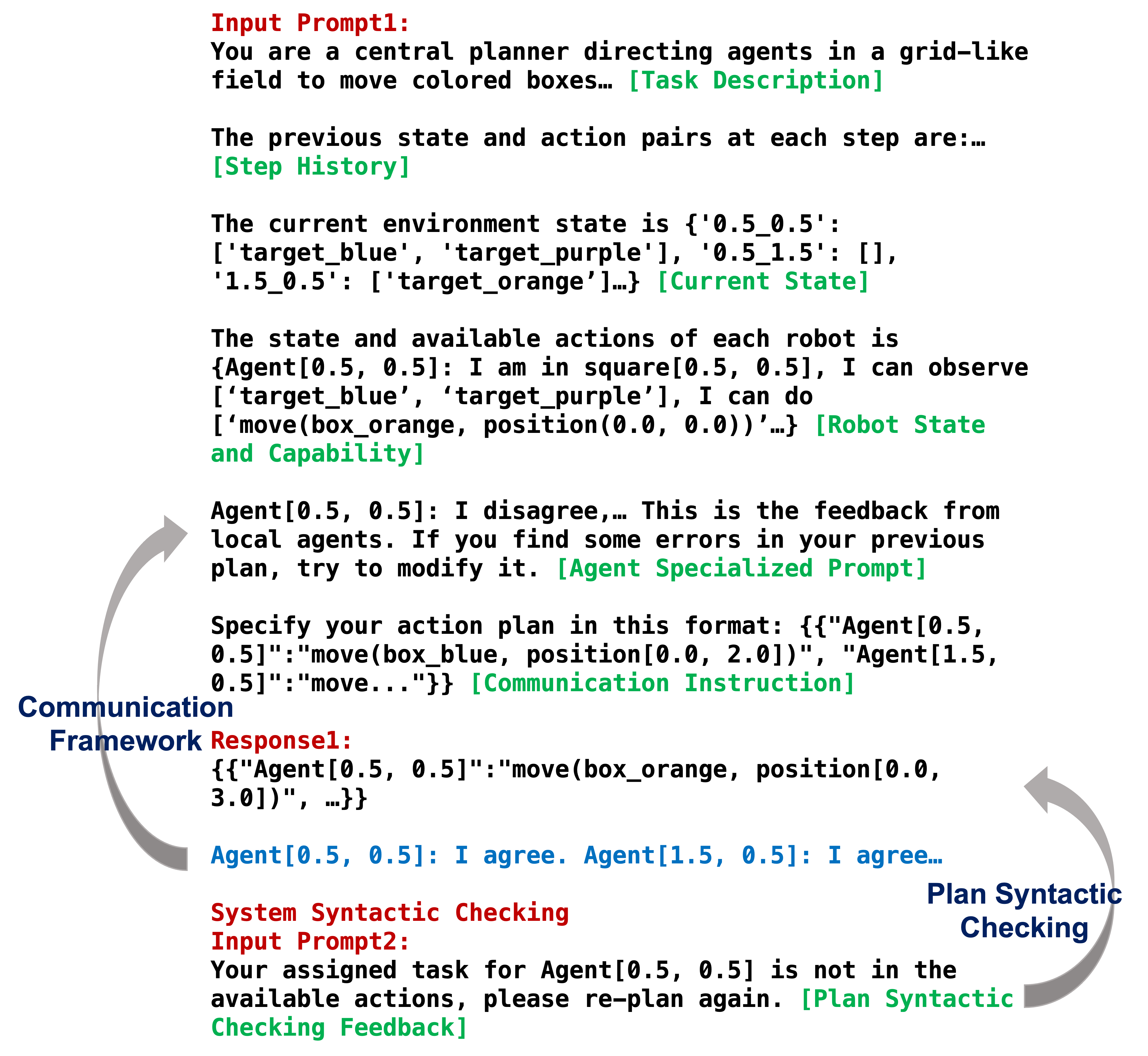}
   \caption{
Simplified prompt example of the HMAS-2 central agent. The generated 'Response1' is sent to local agents for feedback. Once the central-local iteration terminates, the output plan is checked for syntactic correctness.}
   \label{fig:Prompt example simplified}
\end{figure}

\section{METHODS}
Given the goal and the current environment state as text, the LLM agents engage in dialogue per the communication framework (Section \ref{subsection: Communication Frameworks for Sub-task Plan}) in order to generate an initial set of actions for the robots to take. Before execution, this action set is checked by an external rules-based verifier for syntax errors; any errors are provided as feedback to re-prompt for correction. Given a syntactically correct set of actions, the robots then execute those actions in the environment, resulting in a new environment state. We show examples of the prompt structure for the HMAS-1 and HMAS-2 approaches in Figure \ref{fig:Prompt example simplified version2} and Figure \ref{fig:Prompt example simplified} respectively. We describe the main components of the initial prompt in the next subsection.

\subsection{Main Components of LLM Prompt}
\label{subsection: Main Components of LLM Prompt}

We use the same basic structure to prompt each LLM agent, but the specifics of the prompt depend on the individual agent's role. The prompt structure consists of the following main components:
\begin{itemize}
    \item \textbf{Task Description}: the requirements and constraints of the task for the multi-robot system to accomplish.
    \item \textbf{Step History}: the history of dialogue, environment states, and actions from previous steps in the iterative planning process. We describe this in more detail in Section \ref{subsection: round history}.
    \item \textbf{Current State}: the objects in the environment (boxes) and their properties (position and volume).
    \item \textbf{Robot State \& Capability}: the capabilities (available actions) of each robot and their current locations. This prompt part is synthesized by our pre-defined functions. The available actions include possible collisions; it is the responsibility of the planner to find safe plans.
    \item \textbf{Agent Specialized Prompt}: the prompt for each local agent emphasizes its own state and indicates the responses and initial plans of the other agents. For frameworks with a central agent, the prompt for that agent includes feedback from the local agents. Further, each agent is provided a persona.
    \item \textbf{Communication Instruction}: the instruction for how to respond to other agents \& how to format the output.
    \item \textbf{Plan Syntactic Checking Feedback}: (optional) explanation of syntax errors in the generated output. The syntactic checking ensures that the output is formatted correctly and uses available actions.
\end{itemize}

\subsection{Communication Frameworks for Sub-task Plan}
\label{subsection: Communication Frameworks for Sub-task Plan}
We compare the four LLM-based multi-robot planning frameworks shown in Figure \ref{fig:All frameworks}. The Decentralized Multi-agent System framework (DMAS) is shown in Figure \ref{fig:All frameworks}(a) and is the framework used in previous works on LLMs as multi-robot planners \cite{yilun-multi-agent,roco}. Each robot is assigned an LLM planner and another agent to whom it should send its comments. The agents use a turn-taking approach for dialogue, as illustrated. The comments from prior agents in the dialogue are concatenated and included as part of the prompt for the next agent; thus, the prompt length increases over the duration of the dialogue for the current planning iteration. The dialogue ends once the current agent outputs ``EXECUTE" followed by the action for each agent.

The Centralized Multi-agent System framework (CMAS) is shown in Figure \ref{fig:All frameworks}(c). This approach incorporates only a single LLM as a central planner that is responsible for assigning the actions for each robot at each planning iteration.

We propose two Hybrid Multi-agent System frameworks, HMAS-1 (Figure \ref{fig:All frameworks}(b)) \& HMAS-2 (Figure \ref{fig:All frameworks}(d)), that are variants of DMAS and CMAS respectively. In HMAS-1, a central LLM planner proposes an initial set of actions for the current planning iteration that is provided to each of the robots' LLM planners; the robots' LLMs then proceed as done in DMAS. In HMAS-2, a central LLM planner generates an initial set of actions for each robot, as done in CMAS; however, each robot has an LLM agent that checks its assigned action and provides feedback to the central planner. In the case of a local agent disagreeing with its assigned action, the central agent will re-plan. This process repeats until each robot's LLM agrees with its assigned action. Note that in both HMAS-1 and HMAS-2, only the agents that will take an action participate in the dialogue, thus reducing the duration of dialogue and the corresponding number of tokens in the prompts.

\subsection{Step History}
\label{subsection: round history}
Including the full history of the dialogue, environment states, and actions rapidly exhausts the context token budget for the LLM planners, constraining the performances. We therefore compare three approaches in an ablation study of the historical information included in the context: (1) no historical information, (2) only state-action pair history (no dialogue), and (3) the full history. We only include the results for all three approaches in the ablation study; since we found that (2) has the best trade-off between task performance and token efficiency (see Section \ref{subsection: Results}), all other experiments are performed with only state-action pair history. 

\subsection{Token Length Constraint}
We use gpt-4-0613 and gpt-3.5-turbo-0613 in this work, with context token limits of 8192 and 4097, respectively. To make sure the total token length (prompt + response) does not surpass these limits, we employ a sliding context window over the step history part of the prompt; the step history will include as many of the most recent steps as possible without surpassing a total prompt length of 3500 tokens.

\begin{figure}[ht]
  \centering
  \includegraphics[width=1.0\linewidth]{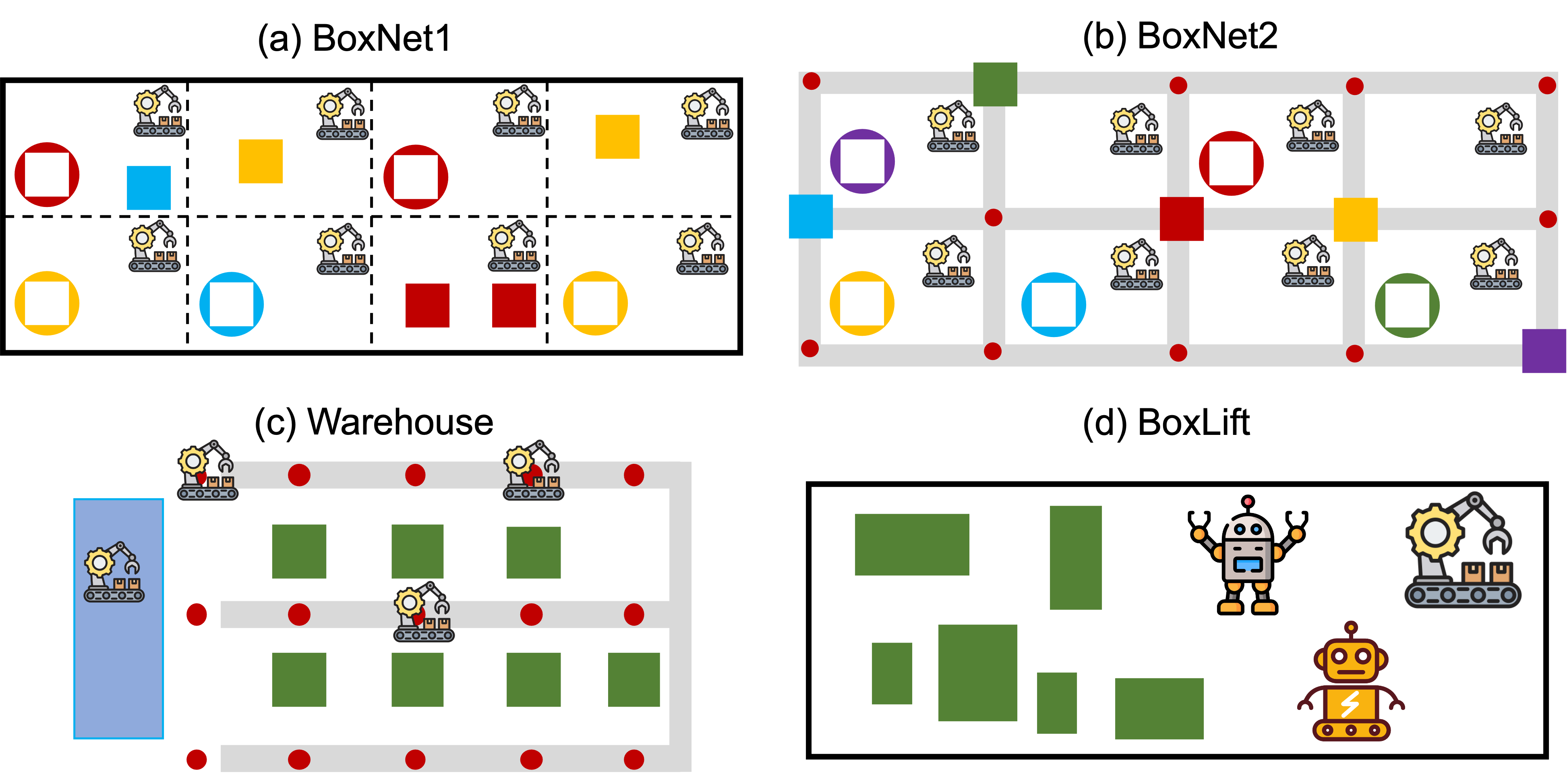}
   \caption{
Four multi-robot task planning environments.}
   \label{fig:All env}
\end{figure}

\section{EXPERIMENTS}
\subsection{Testing Environments}
\label{subsection: Testing Environments}
To compare the four different LLM-based planning frameworks, we design four multi-robot task planning environments inspired by a warehouse setting. In order to evaluate how these frameworks scale to many robots, for BoxNet1 and BoxNet2, we run trials of 4, 8, 16, and 32 robots. For Warehouse and BoxLift, we run trials 4, 6, 8, and 10 robots. For each number of robots in each environment, we perform 10 trials with varied initial conditions.

We track whether each trial resulted in task execution. A task is considered a failure in the following conditions: (1) the dialogue among agents results in a context length beyond the token limit, (2) the agents do not reach consensus before a pre-specified limit of dialogue rounds, (3) the syntactic checking iterates beyond a pre-specified limit, (4) the number of planning iterations exceeds a limit before reaching the goal, and (5) the plan results in a collision. Note that only BoxNet2 and Warehouse can have collision.\\
\textbf{BoxNet1}\quad Figure~\ref{fig:All env}(a) shows the BoxNet1 environment. The environment consists of cell regions, robot arms, colored boxes, and colored goal locations (circles) for each box. The goal is to move each box into its associated goal location in the fewest time steps. The robot arms are confined to the cell they occupy. Each arm has three possible actions: (1) move a box within its cell to a neighboring cell, (2) move a box within its cell to a goal location within its cell, and (3) do nothing. We assume no collisions.\\
\textbf{BoxNet2}\quad Figure~\ref{fig:All env}(b) shows the BoxNet2 environment, which is similar to BoxNet1. In this environment, each box can only be moved between cells by being placed at a corner (red circles), and a given corner can only hold one box at a time; we treat placing two or more boxes on the same corner as a collision (and thus task failure). Each arm in this environment has three possible actions: (1) move a box from a corner to a different corner of the cell, (2) move a box from a corner to a goal location within the its cell, and (3) do nothing. The constraint on box movement and possibility of collision makes this scenario more challenging than BoxNet1.\\
\textbf{Warehouse}\quad Figure~\ref{fig:All env}(c) shows the Warehouse environment. In this environment, mobile manipulators are tasked with moving all of the boxes (green) to the target region (blue) in the fewest time steps. Each robot can only move between permissible locations (red) by traveling along the gray paths; in a single time step, a robot cannot move beyond an adjacent permissible location. We treat two robots occupying the same location as a collision, resulting in task failure. A robot can pick up a box only when at a permissible location that is immediately adjacent to it. Each robot has six possible actions: (1) \& (2) move left or right (if a permissible location exists), (3) pick up an adjacent box, (4) place a box in the target region, (5) move from the target region to any of the adjacent permissible locations, and (6) do nothing.\\
\textbf{BoxLift}\quad Figure~\ref{fig:All env}(d) shows the BoxLift environment. In this environment, robots are tasked to lift each box (green) in the fewest time steps. The robots are able to lift different amounts of weight, and the boxes have different sizes and weights. In a single time step, multiple robots can be assigned to lift the same box. The box is lifted if the total capability of the robots lifting is greater than the box's weight. As an additional challenge, the LLM agents are only able to observe the size of each box, not their weight. This is meant to simulate real situations in which box size roughly correlates with weight. The size and weight of each box is roughly proportional, but we introduce some variability. The LLM agents are provided feedback about whether or not the box was successfully lifted.

\subsection{Metrics}
\label{subsection: Metrics} 
To measure how well each framework is able to plan, we report the average task success rate and average number of steps per plan. To measure the token efficiency and API usage, we also report the average number of tokens used per plan and the average number of API calls per plan. The average number of steps per plan, the average number of tokens per plan, and the average number of API calls per plan only include plans that were successful. We therefore report normalized values for those three metrics. Let $\mathcal{M}$ be the set of values for a given metric $M$, e.g. average API calls, such that $m_i \in \mathcal{M}$ is the value of metric $M$ for the $i^{th}$ framework. Let $\hat{\mathcal{M}}$ be the set of values for the normalized metric $\hat{M}$ such that the normalized metric value $\hat{m}_i \in \hat{\mathcal{M}}$ for the $i^{th}$ framework is:

\begin{equation}
    \hat{m}_i = \frac{m_i}{min(\mathcal{M})}
\end{equation}

The best value for the normalized metrics is 1.0. The framework that performs best for one of those metrics will thus have a value of 1.0.
\subsection{Results}
\label{subsection: Results}
\begin{figure}[ht]
  \centering
  \includegraphics[width=0.7\linewidth]{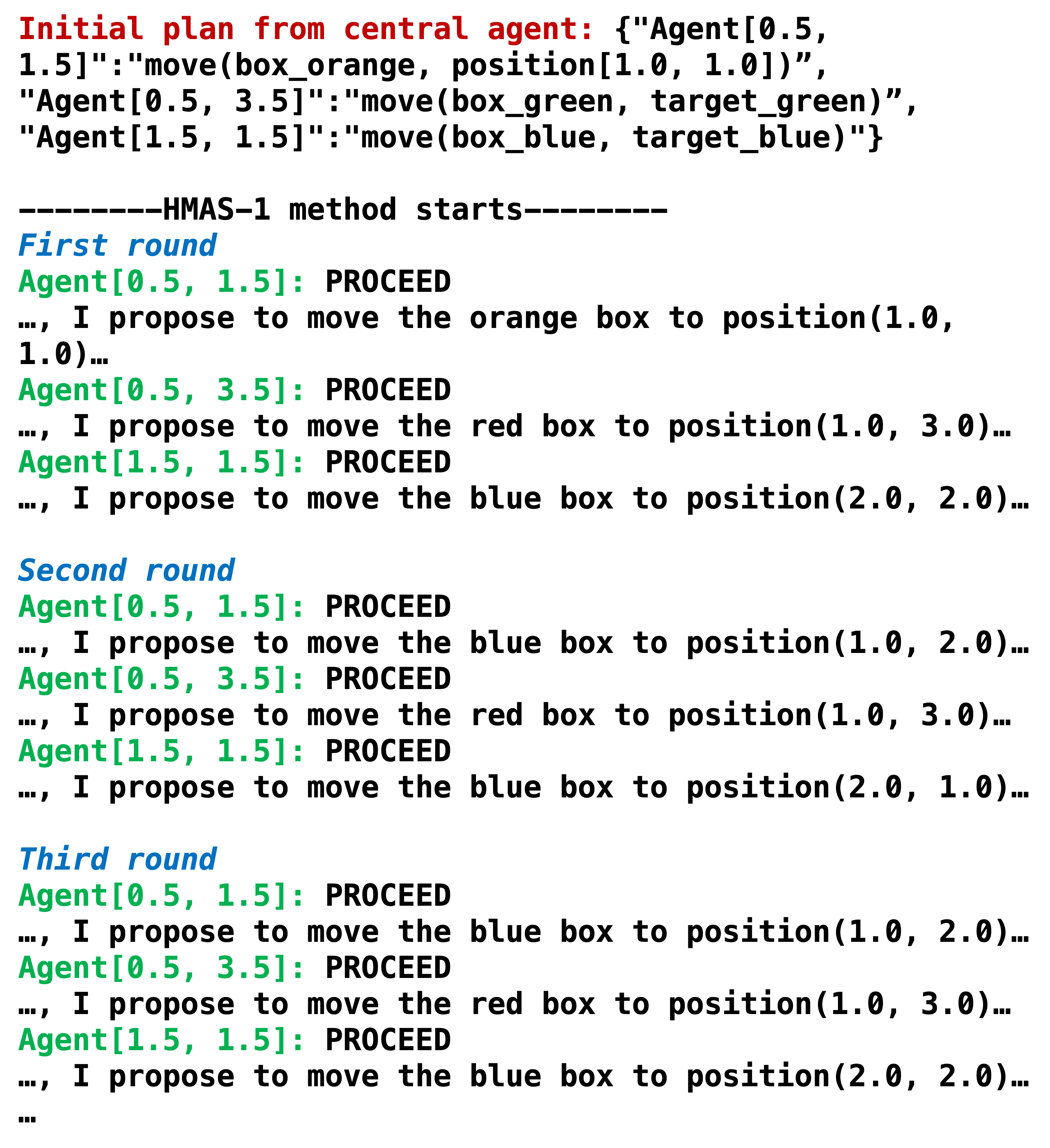}
   \caption{
Simplified communication example of HMAS-1. The local agents hesitate on possible actions, making the dialogue endless.}
   \label{fig:prompt-HMAS1}
\end{figure}

Table~\ref{tab:result_table1} shows the experimental results for the four LLM-based multi-robot planning frameworks.\\
\textbf{Communication Frameworks}\quad We note a few key results. The HMAS-2 framework outputs plans with highest quality since it achieves highest success rates and the fewest actions per plan. The CMAS framework has the fewest API calls and uses the fewest tokens; this is expected as it uses a single LLM and only requires one API call to generate the plan (assuming no syntax errors). The DMAS framework uses the most API calls and tokens, and also has the lowest task success rate. We observe that the LLM agents in DMAS often take many rounds of dialogue per planning step to decide to act, resulting in long dialogues. During the dialogue, the agents often repeat what previous agents have said without contributing anything new; or, agents will repeatedly propose the same action, diluting the context of important information \cite{information_diluted}. HMAS-1, our hybrid variant of DMAS, primes the dialogue with an initial plan from a central LLM planner. This modification significantly improves the performance of the dialgoue that follows. We hypothesize that the initial plan serves as better starting point than DMAS, thus leading to better performance metrics. However, HMAS-2 outperforms HMAS-1 in all metrics. We show one example of HMAS-1 dialogue in Figure~\ref{fig:prompt-HMAS1}. It shows that the LLM agents can get stuck on their proposed action, leading to inefficient dialogue.

\begin{table*}[]
\vspace{2mm}
\caption{Evaluation results on four tasks. We report average success rates ($\uparrow$) over 40 runs across 4 agent numbers per task, average number of steps per plan ($\downarrow$), average API calls ($\downarrow$), and average number of tokens across all runs ($\downarrow$). Apart from success rate, the later three values are normalized over successful runs.}
\centering
\begin{tabular}{|ll|cccc|cc|cc|}
\hline
&   & DMAS & HMAS-1 & CMAS & HMAS-2
    & \begin{tabular}[c]{@{}c@{}}HMAS-2\\ w/o History\end{tabular} & \begin{tabular}[c]{@{}c@{}}HMAS-2\\ w/ All History\end{tabular}
    & \begin{tabular}[c]{@{}c@{}}CMAS\\ (GPT-3)\end{tabular} & \begin{tabular}[c]{@{}c@{}}HMAS-2\\ (GPT-3)\end{tabular} \\ \hline
\multicolumn{1}{|l|}{\multirow{4}{*}{BoxNet 1}}
    & Success & 25.0\% & 52.5\% & 75.0\% & \textbf{82.5\%} & 77.5\% & 75.0\% & 12.5\% & 27.5\% \\
    \multicolumn{1}{|l|}{} & Steps & 2.25 & 1.58 & 1.92 & \textbf{1.33} & 1.45 & 1.34 & 2.76 & 1.74 \\
    \multicolumn{1}{|l|}{} & API Calls & 16.00 & 5.06 & \textbf{1.89} & 4.62 & 4.61 & 4.41 & 3.21 & 6.33 \\
    \multicolumn{1}{|l|}{} & Tokens & 48.56 & 13.80 & \textbf{4.09} & 9.00 & 4.18 & 8.95 & 6.73 & 10.02 \\ \hline
\multicolumn{1}{|l|}{\multirow{4}{*}{BoxNet 2}}
    & Success & 0.0\% & 7.5\% & 27.5\% & \textbf{57.5\%} & 27.5\% & 32.5\% & 0.0\% & 5.0\% \\
    \multicolumn{1}{|l|}{} & Steps & - & 1.38 & 1.22 & \textbf{1.21} & 1.69 & 1.24 & - & 1.29 \\
    \multicolumn{1}{|l|}{} & API Calls & - & 8.24 & \textbf{1.00} & 3.35 & 4.20 & 2.39 & - & 3.97 \\
    \multicolumn{1}{|l|}{} & Tokens & - & 9.32 & \textbf{1.00} & 5.22 & 3.89 & 3.16 & - & 5.83 \\ \hline
\multicolumn{1}{|l|}{\multirow{4}{*}{Warehouse}}
    & Success & 0.0\% & 5.0\% & 15.0\% & \textbf{62.5\%} & 45.0\% & 52.5\% & 0.0\% & 10.0\% \\
    \multicolumn{1}{|l|}{} & Steps & - & 1.62 & 1.73 & \textbf{1.16} & 1.23 & 1.34 & - & 1.19 \\
    \multicolumn{1}{|l|}{} & API Calls & - & 8.52 & \textbf{1.00} & 5.48 & 5.62 & 5.71 & - & 6.33 \\
    \multicolumn{1}{|l|}{} & Tokens & - & 14.02 & \textbf{1.00} & 9.21 & 9.98 & 10.56 & - & 10.09 \\ \hline
\multicolumn{1}{|l|}{\multirow{4}{*}{BoxLift}}
    & Success & 52.5\% & 67.5\% & 90.0\% & \textbf{100.0\%} & 0.0\% & 90.0\% & 12.5\% & 20.0\% \\
    \multicolumn{1}{|l|}{} & Steps & 1.42 & 1.33 & 1.99 & \textbf{1.11} & - & 1.17 & 2.82 & 1.65 \\
    \multicolumn{1}{|l|}{} & API Calls & 37.13 & 18.99 & \textbf{1.00} & 14.90 & - & 15.21 & 1.76 & 18.92 \\
    \multicolumn{1}{|l|}{} & Tokens & 50.34 & 24.54 & \textbf{1.00} & 21.52 & - & 28.02 & 2.04 & 25.42 \\ \hline
\end{tabular}
\label{tab:result_table1}
\end{table*}

\begin{figure}[ht]
  \centering
  \includegraphics[width=0.8\linewidth]{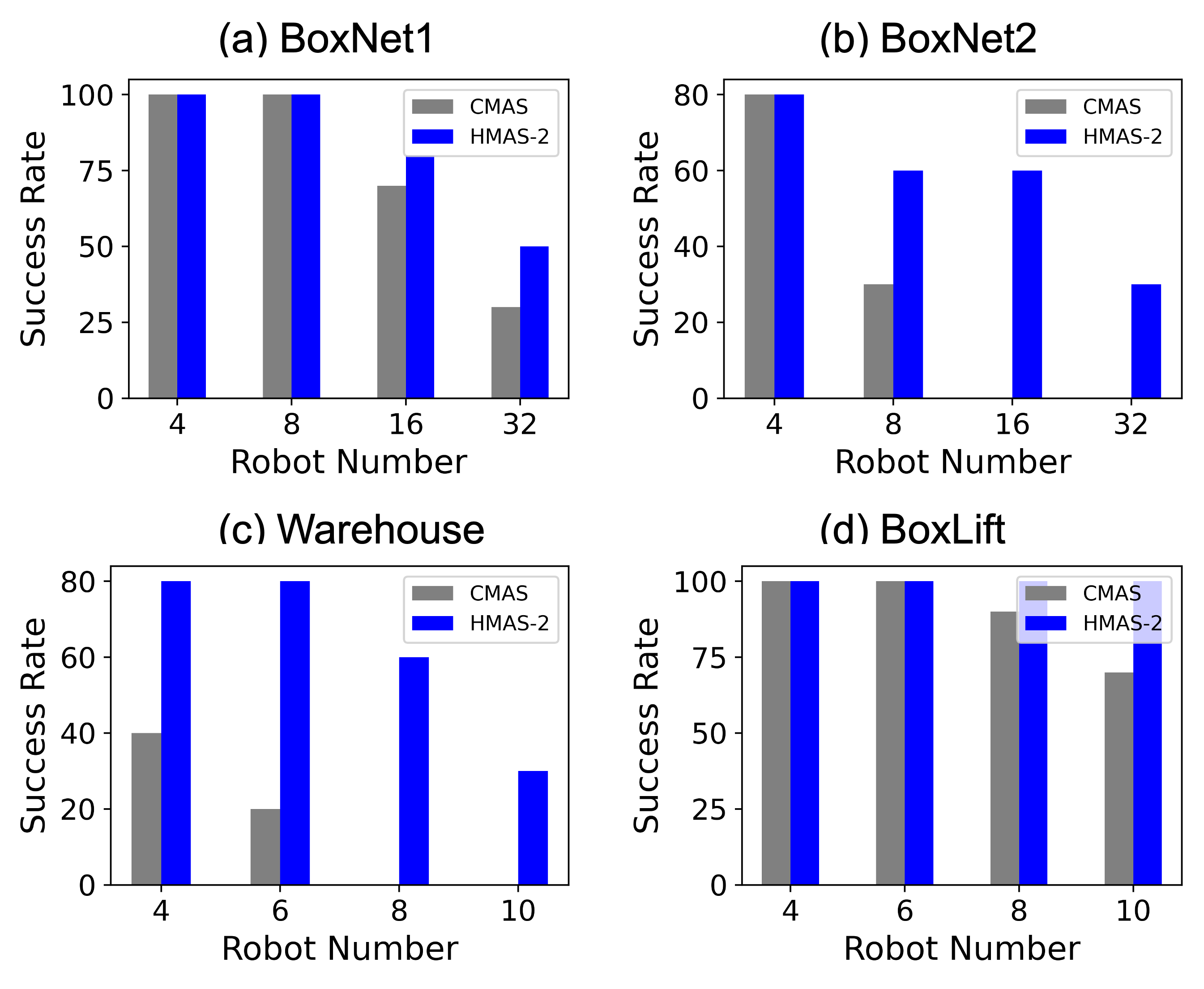}
   \caption{
Success rate vs. robot number for CMAS and HMAS-2 methods in four testing environments.}
   \label{fig:success-agent-num}
\end{figure}

\begin{figure}[ht]
  \centering
  \includegraphics[width=0.8\linewidth]{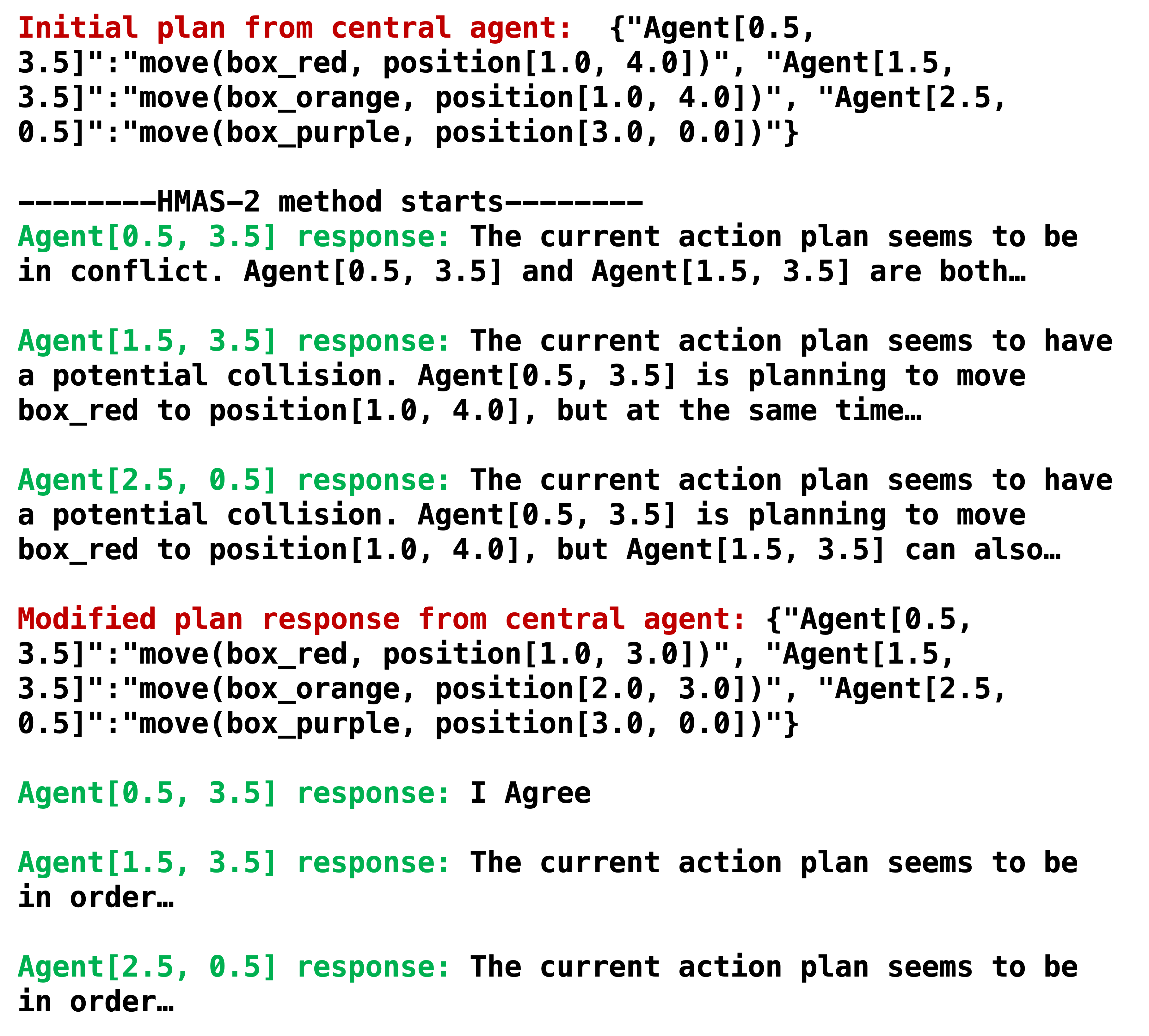}
   \caption{
Simplified communication example of HMAS-2. The local agents detect the collision risk and report it to the central agent.}
   \label{fig:prompt-HMAS2}
\end{figure}

We also report the trend of task success rates as a function of increasing numbers of agents, as shown in Figure~\ref{fig:success-agent-num}. For low numbers of agents, CMAS is competitive with HMAS-2; however, CMAS scales significantly worse to more agents. For more challenging tasks like Warehouse, CMAS performs worse than HMAS-2 for all numbers of agents, indicating that a single central LLM planner tends to generate unreasonable plans for more complex multi-robot task scenarios. Unlike CMAS, HMAS-2 is able to check and correct for errors in plans, such as identifying actions that would result in a collision. Figure~\ref{fig:prompt-HMAS2} shows one example of dialogue correcting the flawed plan via feedback from local LLM agents.\\
\textbf{Step History Method}\quad In Table~\ref{tab:result_table1}, we report the results of HMAS-2 for different step history prompts, as described in Section \ref{subsection: round history}. The framework performs much worse when provided no history of prior actions or dialogue rounds than when provided with the state-action pair history; this is consistent with our intution that the past actions provide useful information for future decisions. The framework performs a bit worse when provided the full history than when provided with the state-action pair history. We hypothesize that this is a result of context dilution \cite{information_diluted} from long dialogue histories.\\
\textbf{GPT-3 Performance}\quad A common trend among pre-trained LLM evaluations is that some capabilities do not emerge until a model reaches sufficient size or is trained on sufficient amounts of quality data \cite{llms-zero-shot-planners}. We report the results of CMAS and HMAS-2 when using GPT-3 as the LLM and find that it performs significantly worse than GPT-4. It is a useful reminder that the quality of LLM-based planners depends on the quality and capability of the underlying LLM.

\begin{figure}[ht]
  \centering
  \includegraphics[width=0.55\linewidth]{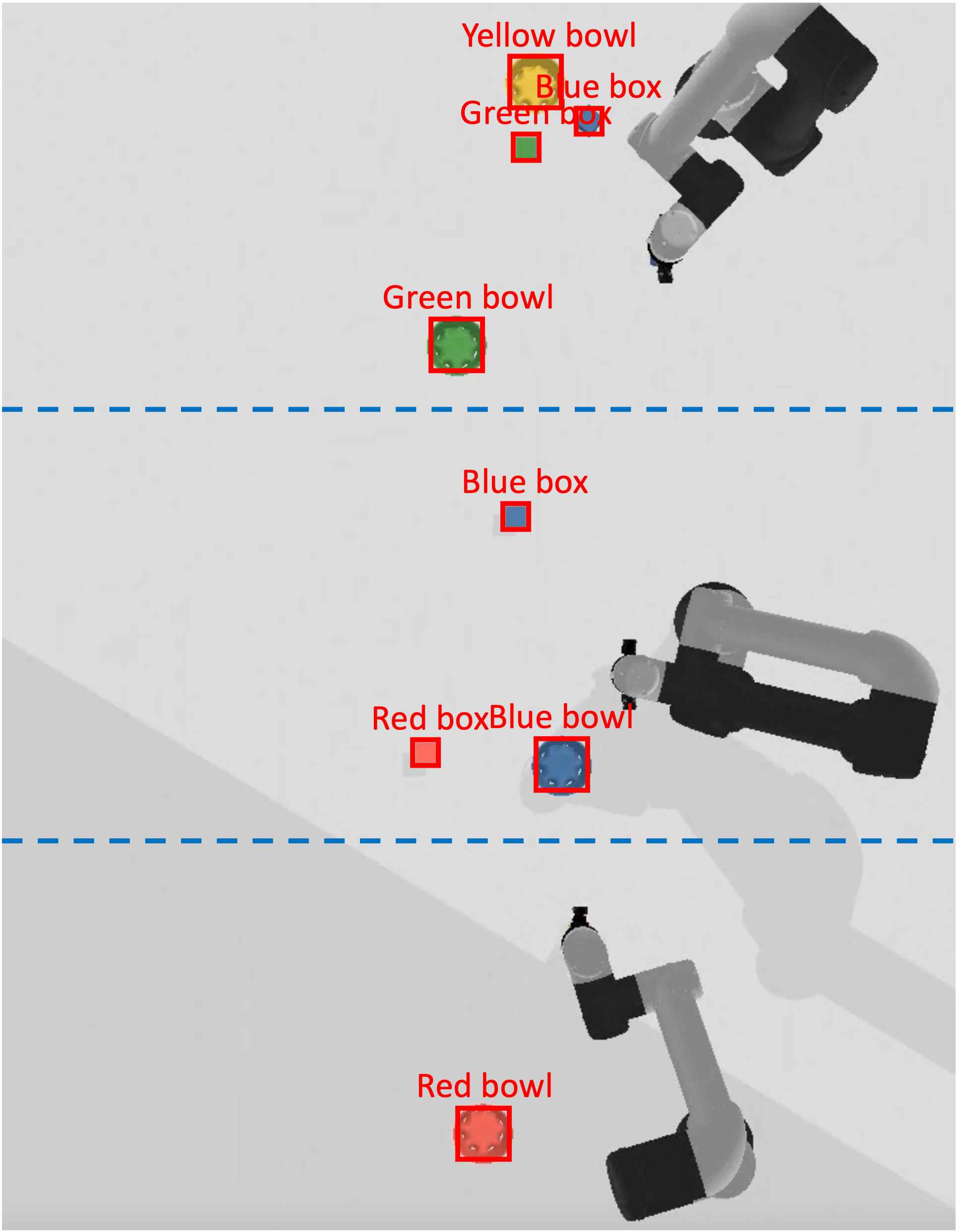}
   \caption{
3D simulation environments: robot arms collaborate to move all the boxes into the same colored bowls. Each robot arm has a limited workspace and can only move within its assigned region (divided by the blue lines).}
   \label{fig:3D simulation env}
\end{figure}
\subsection{3D Simulation}
\label{subsection: 3D Simulation}
In addition to the 2D scenarios for our experiments, we also perform experiments in a 3D environment simulated using Pybullet \cite{pybullet}, as illustrated in Figure~\ref{fig:3D simulation env}. The task and environment are similar to BoxNet1 and BoxNet2. The environment consists of colored boxes, colored bowls, and robot arms. The goal is to move each colored box into its associated bowl of the same color in the fewest actions. Each arm is immobile and confined to actions within its workspace (indicated by the dotted blue lines). Arms can only pick and place boxes that are within its workspace or on the border. Pick and place actions are executed via pre-defined motion primitives. Unlike BoxNet2, boxes can be placed anywhere that is reachable along the boundary, so collisions are possible but unlikely. We again test the scalability to more agents and instantiate the environment with either three or six arms (each has its own workspace). The 3D environment has an additional complexity of using an image-to-text model (ViLD \cite{ViLD}) to provide bounding boxes and text descriptions for each object. Further, the 3D simulation has a richer environment model that permits action execution errors due to dynamical factors (e.g., a box slips out of a gripper) that require re-planning. The iterative nature of the LLM-based planning frameworks in this work naturally handles such instances of replanning.

We do ten runs for each scenario. Table~\ref{tab:result_table2} shows the results of the experiments with three and six agents, respectively. Both CMAS and HMAS-2 achieve 100\% success rates. CMAS used more action steps than HMAS-2 in the six robot situation, consistent with our results from the 2D environments that CMAS performs worse than HMAS-2 in more complex tasks.

\begin{table}[]
\vspace{2mm}
\caption{Evaluation results on 3D simulations. We report average values over 10 runs. Reported numbers of action steps, number of API calls, and number of tokens are normalized values.}
\label{tab:result_table2}
\resizebox{\columnwidth}{!}{
\begin{tabular}{|l|cccc|cccc|}
\hline
    & \multicolumn{4}{c|}{Three Robots} & \multicolumn{4}{c|}{Six Robots} \\ \cline{2-9} 
    & Success & Steps & \begin{tabular}[c]{@{}c@{}}API\\ Calls\end{tabular} & Tokens
        & Success & Steps & \begin{tabular}[c]{@{}c@{}}API\\ Calls\end{tabular} & Tokens \\ \hline
    CMAS & 100\% & 1.0 & 1.0 & 1.0 & 100\% & 1.39 & 1.0 & 1.18 \\
    HMAS-2 & 100\% & 1.03 & 3.45 & 3.28 & 100\% & 1.03 & 2.65 & 1.31 \\ \hline
\end{tabular}
}
\end{table}

\section{RELATED WORK}
\textbf{LLMs for Robotics}\quad A representative set of prior work \cite{chatgpt_for_robot,llms-few-shot-learners,errors-are-useful-prompts,llms-zero-shot-reasoners,llms-construct-and-utilize-world-models-for-task-planning} uses LLMs to select actions from pre-defined skill primitives and complete the tasks step by step with texts or codes as the intermediate, such as SayCan \cite{saycan}, Inner Monologue \cite{inner-monologue}, Code-As-Policy \cite{code-as-policies}, and ProgGPT \cite{progprompt}. Regarding to connect task planning and motion planning, prior work such as Text2Motion \cite{text2motion} and AutoTAMP \cite{autotamp} studies integrating LLMs with traditional Task and Motion Planners. Other work explores querying LLMs to output rewards of robot actions so that independent reward-based planners can be connected. The reward formats can be real values \cite{LLM-for-action-rewards,traffic-generation}, temporal logics \cite{NL2TL, autotamp,lang2ltl}, or patterns \cite{pattern-generation-LLM}. The recent two work \cite{yilun-multi-agent,roco} firstly extend LLMs into multi-robot situations, while the robot number is limited to two or three and the scalability of frameworks and step history approaches is not considered. A recent work \cite{sayplan} considers the scalability of LLM-based single-robot planning in broader environments with more objects.\\
\textbf{Dialogues and Debates of LLMs}\quad Outside the Robotics domain, LLM-based multi-agent discussion has shown impressive capability to promote the research in social behaviors \cite{camel-mind-exploration-with-discussion,traffic-generation}, dialogue-based games\cite{dialogue-games}, and software development \cite{software-development}. Recent work shows that discussion among multiple LLM agents can improve factuality and accuracy\cite{improving-factuality-through-MA-discussion,multi-personna-self-discussion}. Prior work focuses more on understanding LLM behaviors or improving the solution for a single question.\\
\textbf{Multi-Robot Collaboration}\quad Multi-robot collaboration has been extensively studied many decades, especially on multi-arm and multi-drone motion planning \cite{1994multi-arm-manipulation,Multi-agent-path-finding,algorithm-based-MAS}. The traditional methods rely on sampling-based methods for trajectory generation \cite{sampling-based-MAS-trajectory}, or formal methods to optimize Task and Motion Planning \cite{TL-for-multi-agent-time-optimization,CBF-for-multi-agent,algorithm-based-MAS}. Recent work also explored learning-based methods as alternatives \cite{RL-for-dual-arm-motion-planning,learning-multi-arm-motion-planner}.

\section{CONCLUSION}
Our work considers the scalability of LLM-based multi-robot task planning for long-horizon tasks to systems with many robots with heterogeneous capabilities. We propose several new frameworks for collaborative LLM dialogue and find that hybrid approaches with both central and local LLM planners produce the most successful plans and scale best to large number of agents. While the experiments focus on warehouse related tasks, the efficient frameworks can generalize to many other multi-robot collaboration situations.

Future work can explore more complex tasks with more hierarchical frameworks of robot groups, e.g., each agent for each specialized robot sub-group. The rapid arising of multi-modal (particularly visual) models can change the communication paradigm for large multi-agent problems, opening up more research topics on agent framework optimization.

\section{ACKNOWLEDGEMENTS} \label{sec:ACKNOWLEDGEMENTS}
This work was supported by ONR under Award N00014-22-1-2478, Army Research Laboratory under DCIST CRA W911NF-17-2-0181, and MIT-IBM Watson AI Lab. This article solely reflects the conclusions of its authors.



\bibliographystyle{IEEEtran.bst}
\bibliography{references}

\begin{thebibliography}{10}
\providecommand{\url}[1]{#1}
\csname url@rmstyle\endcsname
\providecommand{\newblock}{\relax}
\providecommand{\bibinfo}[2]{#2}
\providecommand\BIBentrySTDinterwordspacing{\spaceskip=0pt\relax}
\providecommand\BIBentryALTinterwordstretchfactor{4}
\providecommand\BIBentryALTinterwordspacing{\spaceskip=\fontdimen2\font plus
\BIBentryALTinterwordstretchfactor\fontdimen3\font minus \fontdimen4\font\relax}
\providecommand\BIBforeignlanguage[2]{{%
\expandafter\ifx\csname l@#1\endcsname\relax
\typeout{** WARNING: IEEEtran.bst: No hyphenation pattern has been}%
\typeout{** loaded for the language `#1'. Using the pattern for}%
\typeout{** the default language instead.}%
\else
\language=\csname l@#1\endcsname
\fi
#2}}

\bibitem{Catl-for-multi-agent}
W.~Liu, K.~Leahy, Z.~Serlin, and C.~Belta, ``Robust multi-agent coordination from catl+ specifications,'' in \emph{2023 American Control Conference (ACC)}.\hskip 1em plus 0.5em minus 0.4em\relax IEEE, 2023, pp. 3529--3534.

\bibitem{autotamp}
Y.~Chen, J.~Arkin, Y.~Zhang, N.~Roy, and C.~Fan, ``Autotamp: Autoregressive task and motion planning with llms as translators and checkers,'' \emph{arXiv preprint arXiv:2306.06531}, 2023.

\bibitem{CBF-for-multi-agent}
M.~Cavorsi, B.~Capelli, L.~Sabattini, and S.~Gil, ``Multi-robot adversarial resilience using control barrier functions,'' in \emph{Robotics: Science and Systems}, 2022.

\bibitem{RL-for-multi-agent-coordination}
F.~Zhang, C.~Jia, Y.-C. Li, L.~Yuan, Y.~Yu, and Z.~Zhang, ``Discovering generalizable multi-agent coordination skills from multi-task offline data,'' in \emph{The Eleventh International Conference on Learning Representations}, 2022.

\bibitem{starcraft}
M.~Samvelyan, T.~Rashid, C.~S. De~Witt, G.~Farquhar, N.~Nardelli, T.~G. Rudner, C.-M. Hung, P.~H. Torr, J.~Foerster, and S.~Whiteson, ``The starcraft multi-agent challenge,'' \emph{arXiv preprint arXiv:1902.04043}, 2019.

\bibitem{llms-few-shot-learners}
T.~Brown, B.~Mann, N.~Ryder, M.~Subbiah, J.~D. Kaplan, P.~Dhariwal, A.~Neelakantan, P.~Shyam, G.~Sastry, A.~Askell, \emph{et~al.}, ``Language models are few-shot learners,'' \emph{Advances in neural information processing systems}, vol.~33, pp. 1877--1901, 2020.

\bibitem{llms-zero-shot-reasoners}
\BIBentryALTinterwordspacing
T.~Kojima, S.~S. Gu, M.~Reid, Y.~Matsuo, and Y.~Iwasawa, ``Large language models are zero-shot reasoners,'' in \emph{ICML 2022 Workshop on Knowledge Retrieval and Language Models}, 2022. [Online]. Available: \url{https://openreview.net/forum?id=6p3AuaHAFiN}
\BIBentrySTDinterwordspacing

\bibitem{saycan}
M.~Ahn, A.~Brohan, N.~Brown, Y.~Chebotar, O.~Cortes, B.~David, C.~Finn, K.~Gopalakrishnan, K.~Hausman, A.~Herzog, \emph{et~al.}, ``Do as i can, not as i say: Grounding language in robotic affordances,'' \emph{arXiv preprint arXiv:2204.01691}, 2022.

\bibitem{llms-zero-shot-planners}
W.~Huang, P.~Abbeel, D.~Pathak, and I.~Mordatch, ``Language models as zero-shot planners: Extracting actionable knowledge for embodied agents,'' in \emph{International Conference on Machine Learning}.\hskip 1em plus 0.5em minus 0.4em\relax PMLR, 2022, pp. 9118--9147.

\bibitem{yilun-multi-agent}
H.~Zhang, W.~Du, J.~Shan, Q.~Zhou, Y.~Du, J.~B. Tenenbaum, T.~Shu, and C.~Gan, ``Building cooperative embodied agents modularly with large language models,'' \emph{arXiv preprint arXiv:2307.02485}, 2023.

\bibitem{roco}
Z.~Mandi, S.~Jain, and S.~Song, ``Roco: Dialectic multi-robot collaboration with large language models,'' \emph{arXiv preprint arXiv:2307.04738}, 2023.

\bibitem{information_diluted}
N.~F. Liu, K.~Lin, J.~Hewitt, A.~Paranjape, M.~Bevilacqua, F.~Petroni, and P.~Liang, ``Lost in the middle: How language models use long contexts,'' \emph{arXiv preprint arXiv:2307.03172}, 2023.

\bibitem{pybullet}
E.~Coumans and Y.~Bai, ``Pybullet, a python module for physics simulation for games, robotics and machine learning,'' \url{http://pybullet.org}, 2016--2021.

\bibitem{ViLD}
X.~Gu, T.-Y. Lin, W.~Kuo, and Y.~Cui, ``Open-vocabulary object detection via vision and language knowledge distillation,'' \emph{arXiv preprint arXiv:2104.13921}, 2021.

\bibitem{chatgpt_for_robot}
N.~Wake, A.~Kanehira, K.~Sasabuchi, J.~Takamatsu, and K.~Ikeuchi, ``Chatgpt empowered long-step robot control in various environments: A case application,'' \emph{arXiv preprint arXiv:2304.03893}, 2023.

\bibitem{errors-are-useful-prompts}
M.~Skreta, N.~Yoshikawa, S.~Arellano-Rubach, Z.~Ji, L.~B. Kristensen, K.~Darvish, A.~Aspuru-Guzik, F.~Shkurti, and A.~Garg, ``Errors are useful prompts: Instruction guided task programming with verifier-assisted iterative prompting,'' \emph{arXiv preprint arXiv:2303.14100}, 2023.

\bibitem{llms-construct-and-utilize-world-models-for-task-planning}
L.~Guan, K.~Valmeekam, S.~Sreedharan, and S.~Kambhampati, ``Leveraging pre-trained large language models to construct and utilize world models for model-based task planning,'' \emph{arXiv preprint arXiv:2305.14909}, 2023.

\bibitem{inner-monologue}
W.~Huang, F.~Xia, T.~Xiao, H.~Chan, J.~Liang, P.~Florence, A.~Zeng, J.~Tompson, I.~Mordatch, Y.~Chebotar, \emph{et~al.}, ``Inner monologue: Embodied reasoning through planning with language models,'' \emph{arXiv preprint arXiv:2207.05608}, 2022.

\bibitem{code-as-policies}
J.~Liang, W.~Huang, F.~Xia, P.~Xu, K.~Hausman, B.~Ichter, P.~Florence, and A.~Zeng, ``Code as policies: Language model programs for embodied control,'' \emph{arXiv preprint arXiv:2209.07753}, 2022.

\bibitem{progprompt}
\BIBentryALTinterwordspacing
I.~Singh, V.~Blukis, A.~Mousavian, A.~Goyal, D.~Xu, J.~Tremblay, D.~Fox, J.~Thomason, and A.~Garg, ``{ProgPrompt}: Generating situated robot task plans using large language models,'' in \emph{International Conference on Robotics and Automation (ICRA)}, 2023. [Online]. Available: \url{https://arxiv.org/abs/2209.11302}
\BIBentrySTDinterwordspacing

\bibitem{text2motion}
K.~Lin, C.~Agia, T.~Migimatsu, M.~Pavone, and J.~Bohg, ``Text2motion: From natural language instructions to feasible plans,'' \emph{arXiv preprint arXiv:2303.12153}, 2023.

\bibitem{LLM-for-action-rewards}
W.~Yu, N.~Gileadi, C.~Fu, S.~Kirmani, K.-H. Lee, M.~G. Arenas, H.-T.~L. Chiang, T.~Erez, L.~Hasenclever, J.~Humplik, \emph{et~al.}, ``Language to rewards for robotic skill synthesis,'' \emph{arXiv preprint arXiv:2306.08647}, 2023.

\bibitem{traffic-generation}
S.~Tan, B.~Ivanovic, X.~Weng, M.~Pavone, and P.~Kraehenbuehl, ``Language conditioned traffic generation,'' \emph{arXiv preprint arXiv:2307.07947}, 2023.

\bibitem{NL2TL}
Y.~Chen, R.~Gandhi, Y.~Zhang, and C.~Fan, ``Nl2tl: Transforming natural languages to temporal logics using large language models,'' \emph{arXiv preprint arXiv:2305.07766}, 2023.

\bibitem{lang2ltl}
\BIBentryALTinterwordspacing
J.~X. Liu, Z.~Yang, B.~Schornstein, S.~Liang, I.~Idrees, S.~Tellex, and A.~Shah, ``Lang2{LTL}: Translating natural language commands to temporal specification with large language models,'' in \emph{Workshop on Language and Robotics at CoRL 2022}, 2022. [Online]. Available: \url{https://openreview.net/forum?id=VxfjGZzrdn}
\BIBentrySTDinterwordspacing

\bibitem{pattern-generation-LLM}
S.~Mirchandani, F.~Xia, P.~Florence, B.~Ichter, D.~Driess, M.~G. Arenas, K.~Rao, D.~Sadigh, and A.~Zeng, ``Large language models as general pattern machines,'' \emph{arXiv preprint arXiv:2307.04721}, 2023.

\bibitem{sayplan}
K.~Rana, J.~Haviland, S.~Garg, J.~Abou-Chakra, I.~Reid, and N.~Suenderhauf, ``Sayplan: Grounding large language models using 3d scene graphs for scalable task planning,'' \emph{arXiv preprint arXiv:2307.06135}, 2023.

\bibitem{camel-mind-exploration-with-discussion}
G.~Li, H.~A. A.~K. Hammoud, H.~Itani, D.~Khizbullin, and B.~Ghanem, ``Camel: Communicative agents for" mind" exploration of large scale language model society,'' \emph{arXiv preprint arXiv:2303.17760}, 2023.

\bibitem{dialogue-games}
D.~Schlangen, ``Dialogue games for benchmarking language understanding: Motivation, taxonomy, strategy,'' \emph{arXiv preprint arXiv:2304.07007}, 2023.

\bibitem{software-development}
C.~Qian, X.~Cong, C.~Yang, W.~Chen, Y.~Su, J.~Xu, Z.~Liu, and M.~Sun, ``Communicative agents for software development,'' \emph{arXiv preprint arXiv:2307.07924}, 2023.

\bibitem{improving-factuality-through-MA-discussion}
Y.~Du, S.~Li, A.~Torralba, J.~B. Tenenbaum, and I.~Mordatch, ``Improving factuality and reasoning in language models through multiagent debate,'' \emph{arXiv preprint arXiv:2305.14325}, 2023.

\bibitem{multi-personna-self-discussion}
Z.~Wang, S.~Mao, W.~Wu, T.~Ge, F.~Wei, and H.~Ji, ``Unleashing cognitive synergy in large language models: A task-solving agent through multi-persona self-collaboration,'' \emph{arXiv preprint arXiv:2307.05300}, 2023.

\bibitem{1994multi-arm-manipulation}
Y.~Koga and J.-C. Latombe, ``On multi-arm manipulation planning,'' in \emph{Proceedings of the 1994 IEEE International Conference on Robotics and Automation}.\hskip 1em plus 0.5em minus 0.4em\relax IEEE, 1994, pp. 945--952.

\bibitem{Multi-agent-path-finding}
B.~Williams, ``Multi-agent path finding for precedence-constrained goal sequences,'' in \emph{International Joint Conference on Autonomous Agents and Multiagent Systems (AAMAS)}, 2022.

\bibitem{algorithm-based-MAS}
V.~N. Hartmann, A.~Orthey, D.~Driess, O.~S. Oguz, and M.~Toussaint, ``Long-horizon multi-robot rearrangement planning for construction assembly,'' \emph{IEEE Transactions on Robotics}, vol.~39, no.~1, pp. 239--252, 2022.

\bibitem{sampling-based-MAS-trajectory}
S.~Karaman and E.~Frazzoli, ``Sampling-based algorithms for optimal motion planning,'' \emph{The international journal of robotics research}, vol.~30, no.~7, pp. 846--894, 2011.

\bibitem{TL-for-multi-agent-time-optimization}
Z.~Liu, M.~Guo, and Z.~Li, ``Time minimization and online synchronization for multi-agent systems under collaborative temporal tasks,'' \emph{arXiv preprint arXiv:2208.07756}, 2022.

\bibitem{RL-for-dual-arm-motion-planning}
C.-C. Wong, S.-Y. Chien, H.-M. Feng, and H.~Aoyama, ``Motion planning for dual-arm robot based on soft actor-critic,'' \emph{IEEE Access}, vol.~9, pp. 26\,871--26\,885, 2021.

\bibitem{learning-multi-arm-motion-planner}
H.~Ha, J.~Xu, and S.~Song, ``Learning a decentralized multi-arm motion planner,'' \emph{arXiv preprint arXiv:2011.02608}, 2020.

\end{thebibliography}

\end{document}